\documentclass[9pt,conference]{IEEEtran}

\usepackage{times}
\usepackage{epsfig}
\usepackage{graphicx}

\usepackage{multirow}
\usepackage{multicol}
\usepackage{pgfplots}
\usepackage{standalone}

\usepackage{amssymb}
\usepackage{algorithm}
\usepackage{algpseudocode}
\usepackage{color}
\usepackage{bbm}
\usepackage{caption}
\usepackage{subcaption}
\usepackage{standalone}
\usepackage{amsmath}
\def\balph{{\boldsymbol{\alpha}}}
\algnewcommand\algorithmicinput{\textbf{Input:}}
\algnewcommand\INPUT{\item[\algorithmicinput]}
\algnewcommand\algorithmicoutput{\textbf{Output:}}
\algnewcommand\OUTPUT{\item[\algorithmicoutput]}

\ifCLASSINFOpdf

\else

\fi

\begin{document}
%
\title{Regularized Residual Quantization: a multi-layer sparse dictionary learning approach}

\author{
\IEEEauthorblockN{Sohrab Ferdowsi, Slava Voloshynovskiy, Dimche Kostadinov
}

\IEEEauthorblockA{Department of Computer Science, University of Geneva, Switzerland\\
$\lbrace$Sohrab.Ferdowsi, svolos, Dimche.Kostadinov$\rbrace$@unige.ch
}

}
\maketitle


\IEEEpeerreviewmaketitle

\section{Introduction}
Quantizing the residual errors from a previous level of quantization has been considered in signal processing for different applications, e.g., image coding. This problem was extensively studied in the 80's and 90's (e.g., see \cite{480761} and \cite{3776}). However, due to strong over-fitting, its efficiency was limited for more modern applications with larger scales. In practice, it was not feasible to train more than a couple of layers. Particularly at high dimensions, the codebooks learned on a training set were not able to quantize a statistically similar test set.


Inspired by an insight from rate-distortion theory,  we introduce an effective regularization for the framework of Residual Quantization (RQ), making it capable to learn multiple layers of codebooks with many stages. Moreover, the introduced framework effectively deals with high dimensions making it feasible to go beyond patch level processing and deals with entire images. The proposed regularization makes use of the problem of optimal rate allocation for asymptotic case of Gaussian independent sources, which is reviewed next.

\section{Background: Quantization of independent Gaussian sources}
Given $n$ independent Gaussian sources $X_j$'s each with variance $\sigma_j^2$ distributed as  $X_j \!\! \sim \!\! \mathcal{N}(0,\sigma_j^2)$, the optimal rate allocation from the rate-distortion theory is derived for this setup as (Ch. 10 of \cite{CoverThomas200607}):
\begin{equation} \label{eq:water-fill_distortion}
D_j =
\begin{cases}
   \gamma ,& \text{if   } \sigma_j^2 \geqslant \gamma, \\
    \sigma_j^2, & \text{if   } \sigma_j^2 < \gamma,
\end{cases}
\end{equation} 
where $\gamma$ should be chosen to guarantee that $\sum_{j=1}^n D_j = D$. Hence, the optimal codeword variance $\sigma_{C_j}^2$ is soft-thresholding of $\sigma_j^2$:

\begin{equation} \label{eq:Sig2C_waterfill}
\sigma_{C_j}^2 = \Big( \sigma_j^2 - \gamma \Big)^+ = 
\begin{cases}
   \sigma_j^2 - \gamma ,& \text{if   } \sigma_j^2 \geqslant \gamma, \\
    0, & \text{if   } \sigma_j^2 < \gamma.
\end{cases}
\end{equation} 

This means that sources with variances less than $\gamma$ should not be assigned any rate at all. We next incorporate this phenomenon for codebook learning and enforce it as a regularization for the codebook variances. This, in fact, will be an effective way to reduce the gap between the train and test distortion errors. Moreover, the inactivity of the dimensions with variances less than $\gamma$ will also lead to a natural sparsity of codewords, which lowers the computational complexity. 

\section{The proposed approach: RRQ} \label{RRQ}

Instead of the standard K-means used in RQ, we first propose its regularized version and then use it as the building-block for RRQ.

\subsection{\textit{VR-Kmeans} algorithm} \label{sub:VR-Kmeans}
\begin{figure}  [!h]
\begin{center}
\includegraphics[width=0.35\textwidth]{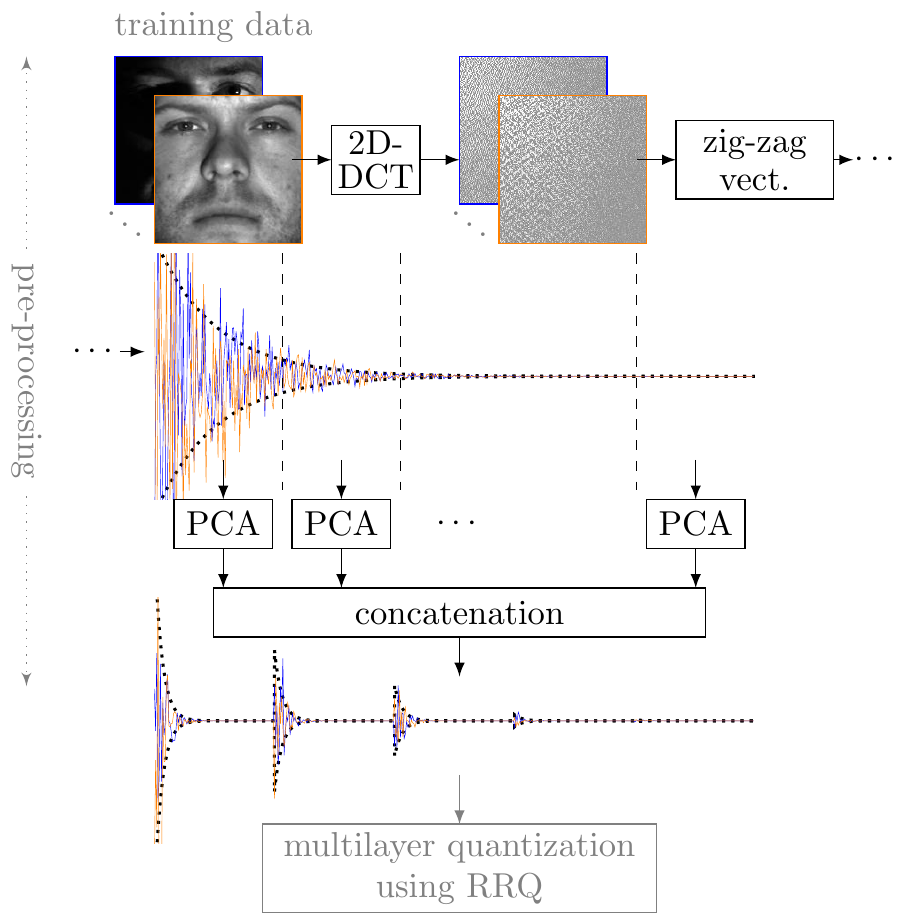}
\end{center}
\caption{A general pre-processing for images required before RRQ. Since global PCA for the full-frame images has $n^2$ parameters, for high-dimensional images it will over-train. Therefore, we propose to use global 2D-DCT which largely de-correlates the images and perform PCA (without dim. reduction) only locally at different sub-bands for further de-correlation. As a result, the data becomes effectively de-correlated with very strong variance-decaying nature, suitable for RRQ.}
   \label{fig:Pre-Processing}
 
\end{figure}

After de-correlating the data points, e.g., using the pre-processing proposed in Fig. \ref{fig:Pre-Processing}, and gathering them in in columns of $\mathrm{X}$ with $\sigma_j^2$ at each dimension, define $\mathrm{S} \triangleq \text{diag}([\sigma_{C_1}^2, \cdots, \sigma_{C_n}^2])$ from Eq. \ref{eq:Sig2C_waterfill}. For codebook $\mathrm{C}$, to regularize only the diagonal elements of $\mathrm{C}\mathrm{C}^T$, define $\mathrm{P}_j$ with all elements as zeros except at $\mathrm{P}_{(j,j)} \!\! = \!\! 1$. We formulate the variance-regularized K-means algorithm with parameter $\lambda$ as:
\begin{equation} \label{eq:VR-Kmeans}
\begin{aligned}
& \underset{{\mathrm{C},\mathrm{A}}}{\text{minimize}}
& & \frac{1}{2} ||\mathrm{X} - \mathrm{C} \mathrm{A}||_F^2 + \frac{1}{2}\lambda || \sum_{j = 1}^n \mathrm{P}_j \mathrm{C} \mathrm{C}^T \mathrm{P}_j - \mathrm{S} ||_F^2,\\
& \text{s.t.}
& & ||\balph_i||_0 = ||\balph_i||_1 = 1. \\
\end{aligned}
\end{equation}  

Like the standard K-means algorithm, we iterate between fixing $\mathrm{C}$ and updating $\mathrm{A}$, and then fixing $\mathrm{A}$ and updating $\mathrm{C}$.  

\textbf{Fix $\mathrm{C}$, update $\mathrm{A}$:} Exactly like the standard K-means.

\textbf{Fix $\mathrm{A}$, update $\mathrm{C}$:} Eq. \ref{eq:VR-Kmeans} can be re-written as:
\begin{equation}  \label{eq:VR-Kmeans-CUpdate2}
\begin{aligned}
 \underset{{\mathrm{C}}}{\text{minimize Tr}}
 & \Big[- \mathrm{X}\mathrm{A}^T\mathrm{C}^T + \frac{1}{2}\mathrm{C}\mathrm{A}\mathrm{A}^T\mathrm{C}^T \\ +& \frac{1}{2}\lambda (\sum_{j = 1}^n \mathrm{P}_j \mathrm{C} \mathrm{C}^T \mathrm{P}_j) (\mathrm{C} \mathrm{C}^T - 2\mathrm{S}) \Big].\\
\end{aligned}
\end{equation}
$\sum_{j = 1}^n \mathrm{P}_j \mathrm{C} \mathrm{C}^T \mathrm{P}_j$, and due to its structure $\mathrm{A}\mathrm{A}^T \triangleq \text{diag}([a_1, \cdots, a_k])$ are diagonal. Therefore Eq. \ref{eq:VR-Kmeans-CUpdate2} will reduce to minimizing independent sub-problems at each (active) dimension:
\begin{equation}  \label{eq:VR-Kmeans-subProblem}
\begin{aligned}
 \underset{{\mathbf{c}(j)}}{\text{minimize}}
 & \Big[ -\mathbf{z}(j)^T\mathbf{c}(j) + \frac{1}{2}\big( a_1c_1(j)^2 + \cdots + a_kc_k(j)^2 \big) \\ +& \frac{1}{2}\lambda ||\mathbf{c}(j)||^2 (||\mathbf{c}(j)||^2 -2\sigma_{C_j}^2) \Big],\\
\end{aligned}
\end{equation}
where $\mathrm{Z} \triangleq \mathrm{X} \mathrm{A}^T = [\mathbf{z}(1), \cdots, \mathbf{z}(n)]^T$. These independent problems can be solved easily using the Newton's algorithm, for which the derivation of the required gradient and Hessian is straightforward.

\subsection{Regularized Residual Quantization (RRQ) algorithm}
For a fixed number of centroids $K^{(l)}$ at layer $l$ and $D_j^{(l-1)}$ the distortion of the previous stage of quantization for each dimension, the RRQ first specifies $\gamma^* = \underset{\gamma}{\text{argmin}} \Big(|\log_2{K^{(l)}} - {\underset{j \in \mathcal{A}_{\gamma}}{\sum} \frac{1}{2}\log_2^+{\frac{D_j^{(l-1)}}{\gamma}}}|  \Big)$ followed by calculation of an active set of dimensions $\mathcal{A}_{\gamma^*}^{(l)} = \lbrace j: 1 \leqslant j \leqslant n | \sigma_j^2 \geqslant \gamma^* \rbrace$. The algorithm then continues with quantizing the residual of stage $l-1$ with the \textit{VR-Kmeans} algorithm described above, until a desired stage $L$ which can be chosen based on distortion constraints or an overall rate budget allowed.
\section{Experiments}
\begin{figure*}  [!h]
\begin{center} 
\subcaptionbox{asymptotic values (if $n \rightarrow \infty$)\label{subfig:VRKmeansDemo_Sig2X}} {\includegraphics[width=0.32\textwidth]{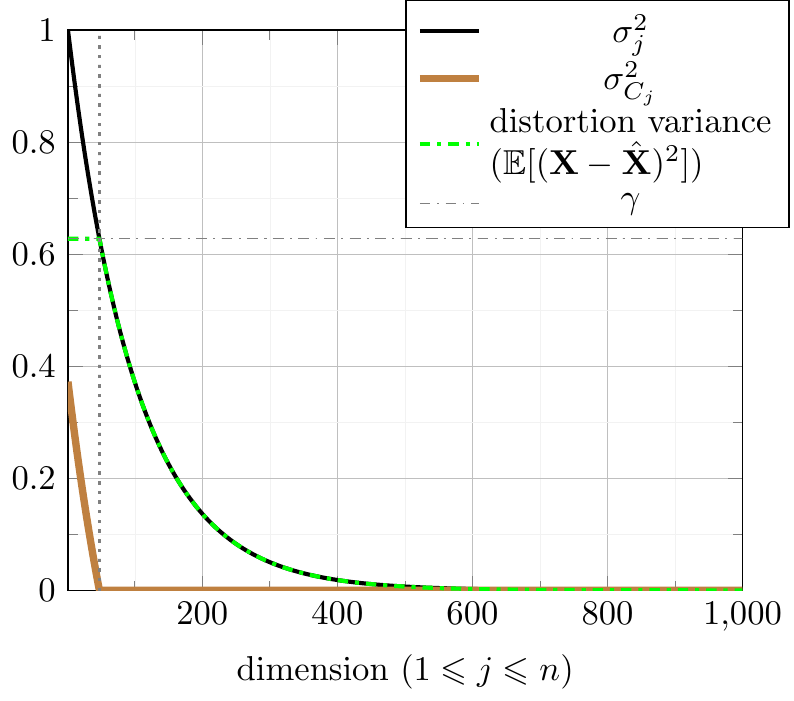}}
\subcaptionbox{codeword variance ($\sigma_{C_j}^2$, log-scale)\label{subfig:VRKmeansDemo_Sig2C}} {\includegraphics[width=0.32\textwidth]{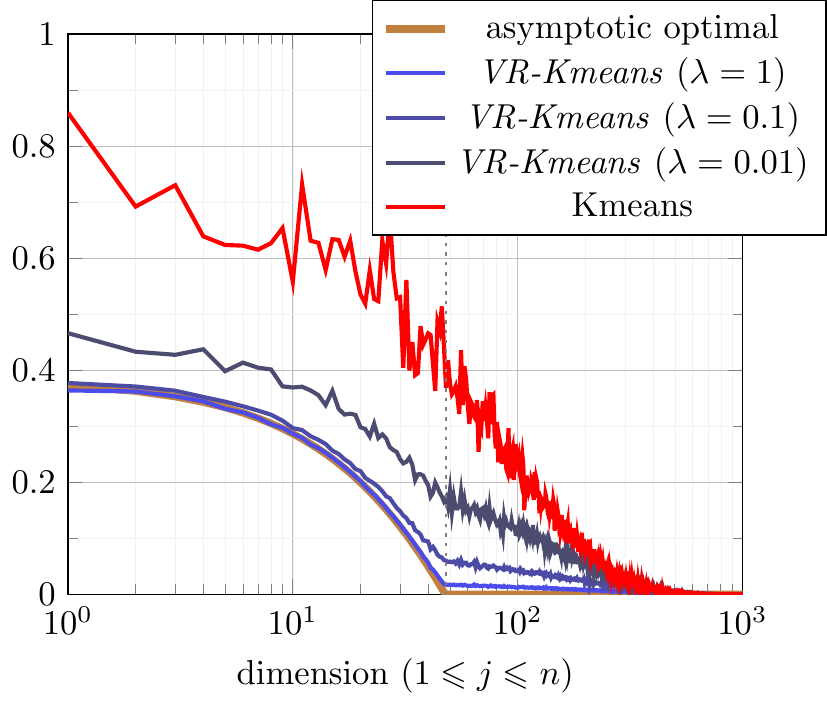}}
\subcaptionbox{distortion variance (zoomed)\label{subfig:VRKmeansDemo_Dvect}} {\includegraphics[width=0.32\textwidth]{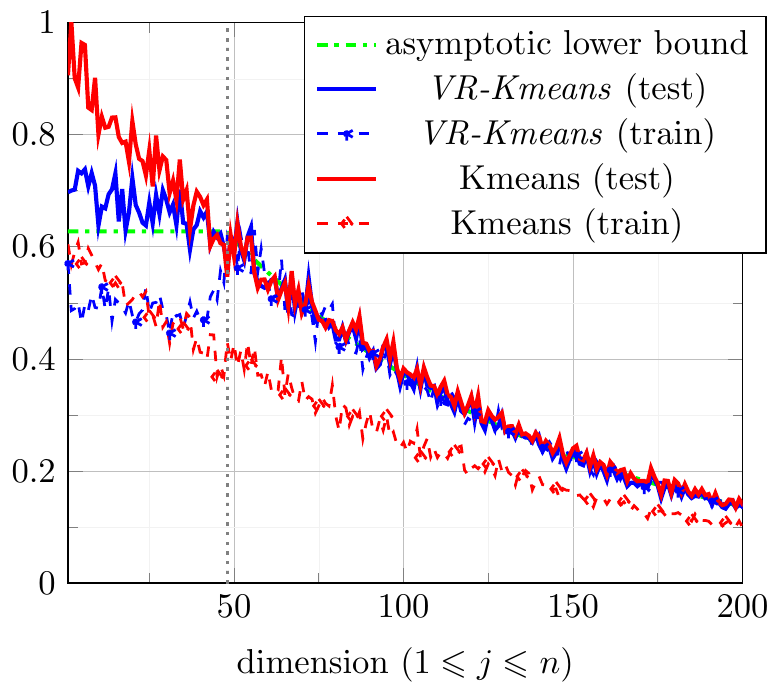}}

   \end{center}

  \caption{Quantization of variance-decaying independent data ($n = 1000$, $K = 256$): (a) asymptotically optimal values (b) Higher values of $\gamma$ enforce the asymptotically optimal values of $\sigma_{C_j}^2$ more strongly. (c) The regularization on \textit{VR-Kmeans} results in less extreme distortion minimization on the training set, but a better performance on the test set.}
   \label{fig:VRKmeansDemo}

   \end{figure*}

Fig. \ref{fig:VRKmeansDemo} and Table \ref{table:VRKmeansDEMO} compare the performance of \textit{VR-Kmeans} with K-means in quantization of high-dimensional variance-decaying independent data. In fact, in many practical cases, the correlated data behaves similarly after an energy-compacting and de-correlating transform. As is seen in this figure, the \textit{VR-Kmeans} regularizes the variance resulting in a reduced train-test distortion gap.

\begin{figure}  

\begin{center}

\subcaptionbox{original\label{subfig:a}} {\includegraphics[width=0.12\textwidth]{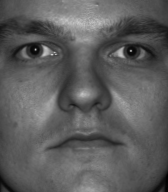}} 
\subcaptionbox{bicubic ($8\uparrow$)\label{subfig:b}} {\includegraphics[width=0.12\textwidth]{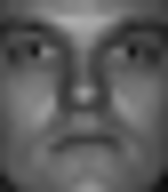}} 
\subcaptionbox{RRQ \label{subfig:c}} {\includegraphics[width=0.12\textwidth]{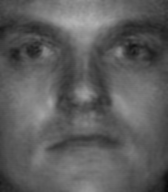}} \\
 \subcaptionbox{original\label{subfig:d}} {\includegraphics[width=0.12\textwidth]{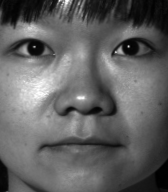}} 
\subcaptionbox{bicubic ($8 \uparrow$)\label{subfig:e}} {\includegraphics[width=0.12\textwidth]{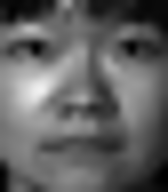}} 
\subcaptionbox{RRQ \label{subfig:f}} {\includegraphics[width=0.12\textwidth]{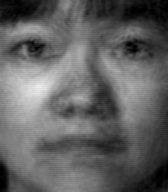}}

\end{center}

\caption{Super-resolution using RRQ on the \textit{CroppedYale} \cite{KCLee05} set with $L = 100$ layers with $K = 256$ centroids each: After pre-processing as proposed in Fig. \ref{fig:Pre-Processing}, a randomly chosen train set is quantized with RRQ. Images from a test set are down-sampled and again up-sampled with bi-cubic interpolation and then quantized and reconstructed by the codebooks learned from the training set. Since the data has strong variance-decaying nature at the initial layers, the trained codewords are very sparse. As the number of layers increases, the data becomes more \textit{i.i.d.} and less structured and hence less sparsity in the codewords.}
   \label{fig:Superresolution}
 
 \end{figure}

Fig. \ref{fig:Superresolution} demonstrates the performance of the RRQ in super-resolution of similar images. It is clear from this figure that the high-frequency content lost in down-sampling can be reconstructed from a multi-layer codebook learned from face images with full resolution.

\begin{table} [!h]
\resizebox{0.5\textwidth}{!}{
\begin{tabular}{c|c|c|c|c|c}
       & Kmeans & \shortstack{{random}\\{generation}} & \shortstack{{\textit{VR-Kmeans}}\\{($\lambda=0.1$)}} & \shortstack{{\textit{VR-Kmeans}}\\{($\lambda=10$)}} & \shortstack{{\textit{VR-Kmeans}}\\{($\lambda=1000$)}} \\ \hline
\multicolumn{1}{c|}{\shortstack{{distortion}\\{train}}} & 0.6727 & 0.9393 & 0.8441 & 0.8520 & 0.8568 \\ \hline
\multicolumn{1}{c|}{\shortstack{{distortion}\\{test}}} & 1.0054 & 0.9394 & 0.9413 & \textbf{0.9384} & 0.9390 \\
\end{tabular}
}
\caption{Quantization distortion (normalized) on the train and test sets for K-means, random codeword generation (from $\mathcal{N}(\mathbf{0},S)$) and the \textit{VR-Kmeans} algorithm (average over 5 trails). The theoretically minimum distortion (achieved at $n \rightarrow \infty$) is $0.9185$. Notice that K-means, while achieves the lowest distortion on the training set, fails to quantize the test set. \textit{VR-Kmeans} with proper $\gamma$, on the other hand, performs the best on the test set.}
\label{table:VRKmeansDEMO}

\end{table}

{\small
\bibliographystyle{ieee}
\bibliography{tempBiblio}
}

\end{document}